\documentclass[Afour,sageh,times]{sagej}

\usepackage{moreverb,url}
\usepackage{hyperref}
\usepackage{graphicx,subcaption}
\usepackage{algorithm,algpseudocode}
\algtext*{EndIf}

\newcommand\BibTeX{{\rmfamily B\kern-.05em \textsc{i\kern-.025em b}\kern-.08em
T\kern-.1667em\lower.7ex\hbox{E}\kern-.125emX}}

\def\volumeyear{2019}

\begin{document}

\runninghead{Fitzsimons et al.}

\title{Task-Based Hybrid Shared Control for Training Through Forceful Interaction}

\author{Kathleen Fitzsimons\affilnum{1}, Aleksandra Kalinowska\affilnum{1}, Julius P. Dewald\affilnum{2}, and Todd D. Murphey\affilnum{1}}

\affiliation{\affilnum{1}Department of Mechanical Engineering, Northwestern University, Evanston, IL,USA
\affilnum{2}Physical Therapy and Human Movement Sciences, Northwestern University, Chicago IL, USA}

\corrauth{Kathleen Fitzsimons, Department of Mechanical Engineering, Northwestern University
2145 Sheridan Rd, Evanston, IL 60208,USA}

\email{k-fitzsimons@u.northwestern.edu}

\begin{abstract}
Despite the fact that robotic platforms can provide both consistent practice and objective assessments of users over the course of their training, there are relatively few instances where physical human robot interaction has been significantly more effective than unassisted practice or human-mediated training.
This paper describes a hybrid shared control robot, which enhances task learning through kinesthetic feedback. The assistance assesses user actions using a task-specific evaluation criterion and selectively accepts or rejects them at each time instant. Through two human subject studies (total n=68), we show that this hybrid approach of switching between full transparency and full rejection of user inputs leads to increased skill acquisition and short-term retention compared to unassisted practice. Moreover, we show that the shared control paradigm exhibits features previously shown to promote successful training. It avoids user passivity by only rejecting user actions and allowing failure at the task. It improves performance during assistance, providing meaningful task-specific feedback. It is sensitive to initial skill of the user and behaves as an `assist-as-needed' control scheme---adapting its engagement in real time based on the performance and needs of the user. Unlike other successful algorithms, it does not require explicit modulation of the level of impedance or error amplification during training and it is permissive to a range of strategies because of its evaluation criterion.
We demonstrate that the proposed hybrid shared control paradigm with a task-based minimal intervention criterion significantly enhances task-specific training.
\end{abstract}

\keywords{Physical Human-Robot Interaction, Rehabilitation Robotics, Human Performance Augmentation}

\maketitle

\section{Introduction}
Approaches to designing kinesthetic feedback for robotic training platforms lie on a spectrum from antagonistic and resistive strategies that are dynamically updated based on user performance to passive assistive strategies in which users have a consistent guide during training. Training regimens at either end of the spectrum have been shown to be appropriate depending on the type and relative difficulty of the task. Passive assistance in the form of virtual fixtures~\citep{rosenberg1993} or record and replay strategies can provide task-relevant feedback to users by demonstrating correct movements. However, this type of guidance may not engage or challenge users because it does not dynamically adapt to different users or changes in user performance. Training in which errors are amplified rather than reduced by guidance has been effective in inducing adaptations in healthy and impaired individuals~\citep{patton2006b} during quasistatic reaching, but guidance was more effective in a timing-based motor task when individuals were less skilled~\citep{milot2010}. Active assistance or shared control has been introduced as an alternative where the level of assistance or impedance is modulated based on performance heuristics. Though the results of robotic training are mixed, meta-analysis of studies using robotics in therapeutic settings demonstrate small but significant improvements in patient outcomes compared to usual care~\citep{krebs2018}.

Here we present a hybrid shared control paradigm that lies in the middle of that spectrum---it does not resist or aid correct actions but requires user action for task completion. 
The autonomy evaluates user inputs based on criteria that capture how well the current input contributes to task completion. If the filtering criterion is met, the controller is transparent to the user. When the criterion is not met, the robot physically rejects the user input, providing feedback but not guidance. 
Rather than adjusting the relative contributions of the robot and human on a continuum based on heuristics over past performance of the user, we hypothesize that using an evaluation criterion to instantaneously switch between full user control and full rejection of user actions by the autonomy is sufficient to improve user performance, adapt to user skill, and ultimately enhance learning of a task.

The user input is evaluated at each time instant, using methods from model predictive control, which allows us to avoid prescribing a desired trajectory over time. 
This enables users to try different task completion strategies, to make errors, and to fail---all of which are critical to learning~\citep{thoroughman2000,lewek2009,koenig2016}. Additionally,  the fact that we choose to only reject user input rather than replacing user input means that users must engage in the task actively to achieve success. The results of two user studies demonstrates that the controller-filter also adapts to the initial skill of the users, and adjusts the level of assistance based on current user performance much like an assist-as-needed controller. It does this without any pre-training assessments of the user's initial skill and without evaluating the overall performance of the subject within the current trial or any preceding trials. We find that this form of hybrid shared control is an effective training tool for both improving skill acquisition and retention of skill one week post-training.

In this paper we show that a hybrid approach to switching between full user autonomy and full rejection of user inputs is an effective way to enhance learning through forceful interaction with a robot. Furthermore, we show, through two user studies, that the task-based switching control leads to improved subject performance while the assistance is engaged, decreased intervention for highly skilled users, and assistance that increases when subject performance is poor and becomes more transparent when subjects perform well.

The paper is organized as follows. First, we review relevant work in robotic training in Section~\ref{sec:background} and our prior work in Section~\ref{sec:priorwork}. We introduce the hyrbid shared control algorithm in Section~\ref{sec:filter} and discuss the task-based criteria used to assess user inputs in Section~\ref{sec:criteria}. The experimental platform and protocol is discussed in Sections \ref{nact3d} and \ref{sec:exp}, respectively. Experimental results of two user studies are given in Section \ref{sec:results}---discussing the training effect in Section~\ref{sec:train} and the relevant features in Section~\ref{sec:assist}-\ref{sec:aan}. Finally, a discussion of the results and their implications for future work is given in Section \ref{sec:disc}. 

\section{Relevant Background}\label{sec:background}

Using robotics in training provides a platform for consistent, high intensity repetitions that are not limited by the time the coach or therapist has available. In rehabilitation settings, specifically, devices can provide support and safety---reducing the physical and cognitive load of the caregiver. Patients who receive additional therapy with robotics often have improved clinical outcomes compared to patients receiving the standard of care~\citep{lum2002,volpe2005,reinkensmeyer2004,krebs2007,squeri2014}. Furthermore, robotics can quantitatively assess users~\citep{stienen2011} and have the potential to systematically tailor the interaction to the user's skill or level of impairment.
As a result, there is interest in facilitating training and rehabilitation through forceful interaction between robots and humans. 

Numerous devices and control strategies have been developed to support physical human robot interaction (pHRI) and modulate it based on principles of motor learning. 
Despite the development of novel hardware and software to facilitate pHRI for training and therapy, there are relatively few instances where robotics have been used to significantly improve learning outcomes. Gains are often modest~\citep{prange2009,mehrholz2013} or equivalent to a similar amount of human-mediated training~\citep{lo2010,veerbeek2014,dobkin2012}. 
The success of robot-mediated therapy is highly dependent on the principles used to design robotic assistance and the corresponding features of training interfaces, which vary greatly from one implementation to another.

Traditional robotic control techniques have been designed to minimize error with respect to a desired trajectory or produce motions that minimize an objective function consisting of both error and effort components. Early rehabilitation robotics used a recorded trajectory from a human expert or healthy reference and `replayed' it with position controllers~\citep{colombo2000,burgar2000}. 
Alternatively, the reference was generated from an optimal task completion, such as minimum jerk reaching in the upper limb~\citep{hogan1984,flash1985}. 
Robotically assisting subjects to perform these normative movements has led to moderate improvements in training outcomes compared to unassisted practice~\citep{kahn2006,bluteau2008,marchal2008}. 
This type of guidance has been especially effective when the learned task is difficult relative to the subject skill level~\citep{guadagnoli2007} or the subject has a high level of impairment~\citep{cesqui2008}.  However, haptic guidance can actually interfere with learning~\citep{schmidt1992,winstein1994,powell2012} or lead to `slacking' by the user~\citep{reinkensmeyer2007,marchal2009}. 
When learning a task, the central nervous system encodes not only a sequence of joint positions but also a feedback control loop---making motor output necessary to learning~\citep{shadmehr1994}. 
So while it is necessary for robotic trainers to be able to assist subjects in completing the task, especially when subjects have limited ability or skill, too much support---leading to user passivity---is not conducive to learning. 

Rather than assisting subjects with task completion, some training paradigms act antagonistically to task goals, making aspects of the task more difficult and allowing failure.  
For instance, robotics have been used to introduce random noise-based disturbances into training. Supported by studies demonstrating that mistakes or errors actually enhance learning~\citep{thoroughman2000}, training with this approach has been show to improve training outcomes compared to progressive guidance strategies and unassisted practice~\citep{lee2010}. Perturbation-based training could also improve the robustness of robot-mediated training---in human-robot teaming training with perturbations led to increased performance across task variants~\citep{ramakrishnan2017}. Alternatively, control strategies that explicitly amplify errors have been developed and have also been shown to improve motor learning in the upper limb~\citep{emken2005,patton2006b,emken2007}, though the effects may be transient or may not generalize to other similar tasks~\citep{patton2006}. 
Interestingly, error amplification is most effective when the users are not novices~\citep{milot2010}, suggesting that this antagonistic strategy is not appropriate for unskilled or highly impaired individuals. 
Finally, another approach is to allow users to  make errors rather than enhancing them explicitly. 
Simply enabling kinematic variability has proved to be more effective than enforcing strict repetitive movement patterns~\citep{lewek2009}. 
As a result, impedance-based shared control has been widely adopted in pHRI to increase kinematic variability and allow users to make errors~\citep{koenig2016}.

While shared control approaches are often implemented to augment user inputs such that task performance is optimized~\citep{dragan2013}, it does not necessarily improve training outcomes~\citep{omalley2006}. The efficacy of blending control signals of a human expert~\citep{khademian2011} or robotic teacher~\citep{perez2016, rakita2018} with students through shared control varies depending on the task and mode of assistance~\citep{powell2012}. 
Generally, shared control for training is considered most effective when the robot provides only as much assistance as is necessary based on estimates of user intent~\citep{li2003,yu2005}, motor contribution~\citep{riener2005}, or other performance heuristics.

Assist-as-needed control schemes are implemented by dynamically updating the relative contributions of the robot and human. Updates to the relative contributions are made by adjusting the gains of an impedance controller based on measured outcomes~\citep{krebs2003,pehlivan2016}, introducing forgetting factors that adjust robot effort according to a schedule~\citep{wolbrecht2008,emken2008}, or implementing a repulsive potential field at the boundary of a virtual tunnel around a desired path~\citep{duschau2010}. 

Numerous implementations of assist-as-needed controllers have been developed for robots that support gait rehabilitation in exoskeletons~\citep{duschau2010}, provide end-point guidance for upper limb tasks~\citep{ferraro2003}, offer support at anatomical joints in upper limb exoskeletons~\citep{wolbrecht2008}, and enhance sports training~\citep{von2008,rauter2010,marchal2013} with mixed results.

Given that approaches at either end of the assistve/resistive spectrum seem to be effective in some cases and ineffective in other training scenarios, one might ask what features of the interfaces discussed above create conditions conducive to motor learning? 
One idea that is consistent across training strategies is the need for user engagement and active participation~\citep{marchal2009}, often accomplished by modulating the assistive or antagonistic forces based on subject performance trial to trial. 
However, it is still unclear how to best implement real-time modulation.
Literature suggests that it is necessary for platforms to be capable of assisting subjects in completing the desired task, especially when the user is unskilled. Yet, allowing or enhancing errors is critical to learning. 
In this paper, we describe a novel shared control paradigm that, through an initial human subject study, we find to be successful in improving learning. We then explore the features of the shared control paradigm in the context of previous findings as described above.

\section{Prior Work}\label{sec:priorwork}
An algorithm for filtering control inputs was proposed in~\citep{therakis2015b} for noise driven swing-up problems based on the hypothesis that noisy inputs can be a rich source of control authority if filtered in a meaningful task-specific way. This filter was implemented by combining a controller and a filter into a single computational unit that cancels noise samples not driving the system towards a desired control direction. 
\begin{figure}[!t]
\centering
\includegraphics[width=0.9\columnwidth]{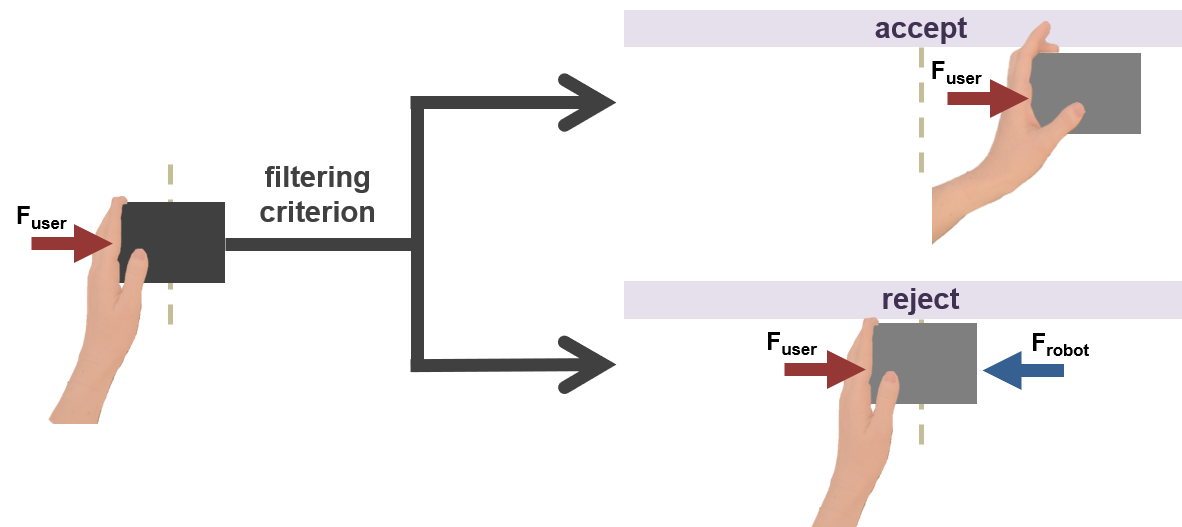}
\caption{Robotic responses of hybrid share control on the example of a hand pushing a mass. The robot filters user input by physically accepting or rejecting it. When a user action is accepted, the robot admits the force. When a user action is not accepted, the robot rejects it by applying an equal and opposite force. }
\label{fig: accept_reject_replace}
\end{figure}
In~\citep{fitzsimons2016} and~\citep{kalinowska2018}, we modified this algorithm to allow for filtering of \emph{user input}.
User inputs were either accepted or rejected based on the criteria described in Sections~\ref{sec:MIGcrit} and \ref{sec:IPcrit}. When they were not accepted, they may be either rejected by the automation (as shown in Figure 1) or replaced with input prescribed by a control policy. 
In the experiments described in this work and our previous work, subject inputs were not replaced---allowing users to fail both allowed us to evaluate the participants' success rate during trials with and without the shared control and to evaluate the training effect of the kinesthetic feedback provided to them.

Previous experiments on a touchscreen platform in \cite{fitzsimons2016} represented an infinite actuation scenario for the filter, since user inputs were able to be completely rejected in software. A haptic stylus (Phantom Omni by Sensable) on the other hand provided kinesthetic feedback, but did not have sufficient power to do more than weakly resist user inputs. We found that both implementations were able to effectively assist subjects in swinging up a cart-pendulum system compared to their baseline performance. The touchscreen platform indicated significantly higher success rates and lower time to success for the swing-up task. Although the assistance mode on the haptic platform did increase the success rate, there was no significant difference in time to success between the baseline and the assistance mode. This was likely due to the fact that the haptic interface did not generate enough force to strictly enforce the filter's acceptance criterion.

Therefore, we realized the mechanical filter on a higher power robotic system described in Section~\ref{nact3d}. Preliminary results of this work  have been discussed in~\citep{kalinowska2018}, where we noted a modest training effect compared to controls with unassisted practice as well as a low, but significant correlation between the controller intervention rate and the participant's initial skill level. In this work, we extend these results by evaluating the progression of subject performance over time. We also present results using an alternative acceptance criterion and assess the skill retention of the trained group after one week.

\section{Methods}
\subsection{Hybrid Shared Control}\label{sec:filter}
The hybrid shared control algorithm works as follows. Given a system and an operator, assume that a user input is measured every $t_s$ seconds. The user input is assessed based on one of the acceptance criterion described by (\ref{eq:migcrit}) or (\ref{inner_product})---roughly asking whether the user understands the task goal or an optimal control strategy for task completion. When the acceptance criterion is met, if the magnitude of the user command is within the allowed limits, the command is applied to the system. Otherwise, saturation may be applied.\endnote{Saturation limits may correspond to physical constraints e.g. angle or torque/force limits etc.} On the contrary, if the criterion is not met, one of two alternatives can be followed: a) the system input can be set equal to zero (user command is ``rejected") or b) the system input can be set equal to the nominal control value. The latter case would result in potentially never-failing interfaces, serving both training and safety purposes. Note that in our experimental setup we followed the first approach; the rationale behind this choice is that being allowed to fail in the task should provide clear indications as to whether the filtering algorithm has any effect on performance. When inputs were rejected in these experiments, a force equal and opposite to the force of the user is exerted at the end-effector. This results in the interface being transparent when user inputs are accepted or velocity being held constant when inputs are rejected.
This process is illustrated in Algorithm~\ref{algorithm}.

\begin{algorithm}
     \caption{Hybrid shared control algorithm}
     \label{algorithm}  
     \hrule  
     \vspace{2 mm}
     Initialize current time $t_0$, sampling time $t_s$, time horizon length $T$, final time $t_f$, input saturation $u_{sat}$ and angle tolerance  $\gamma$.   
     \hrule
      \vspace{2 mm}
\begin{algorithmic}[1]       
\While{$t_0<t_f$}
\State Infer user input $u_{user}$ from sensor data
\State Calculate the quantities in eq.~\ref{eq:migcrit} or \ref{inner_product} for time $T$.
\If{Filter Criterion is True}  
    \If{$|u_{user}| < u_{sat}$\label{saturation}}
		\State Use $u_{user}$ as current input, $u_{curr} = u_{user}$
	 \Else
	   \State Apply saturated user input $u_{curr} = u_{sat}$
	 \EndIf
\Else
     \State Completely ``reject" $u_{user}$ $(u_{curr} = 0)$
\EndIf
\State Apply $u_{curr}$ for $t\in[t_0, t_0+t_s]$
\State $t_0 = t_0+t_s$
\EndWhile

\end{algorithmic}
\end{algorithm}
\subsection{Acceptance Criteria}\label{sec:criteria}
In this paper, we use two criteria. Both are reasonable interpretations of the hybrid philosophy of shared control. The Mode Insertion Gradient (MIG) assumes the user must be generating \emph{descent directions} while the Optimal Controller Inner Product (OCIP) insists that the user \emph{agrees} with the optimal control.  Because of this difference, MIG is more relevant to assessing how well a person understands a task in the moment, whereas OCIP is more relevant to whether the person is being \emph{taught} by the optimal control solutions we compute.  Naturally these two interpretations have considerable overlap, but in different situations the choice may matter.  For instance, a driver-assist wheelchair  may need to interpret the quality of motion control a person is providing without having an explicit need to train the user and potentially having reason to believe that the user needs flexibility in his/her implementation (leading to MIG being a better choice).  On the other hand, technologies geared toward rehabilitation may want to steer a person's motor control towards a normative set of expected solutions (leading to OCIP). \
The practical consequences of these two interpretations of acceptance is that
in the MIG study the acceptance criteria was met much more frequently and user actions were rejected less often than in the OCIP study.
\subsubsection{Mode Insertion Gradient Criterion.}\label{sec:MIGcrit}
The mode insertion gradient $ \frac{dJ}{d\lambda}$ is most often used in mode scheduling problems to determine the optimal time $\tau$ to insert control modes from a predetermined set~\citep{egerstedt2006,wardi2012,vaseduvan2010, ansari2016, caldwell2016}. In these cases, it gives an estimate of the sensitivity of the cost function to the timing of a switch from one control mode to another. Therefore, a negative MIG at a specific time indicates that a mode switch at that time would decrease the cost compared to not switching modes. Often, the goal is to choose an application time when the MIG is most negative, to optimize the benefit of switching control modes. Here we use the mode insertion gradient as a measure of the sensitivity of the cost to a change from the nominal control, $u_1$, to a particular user input, $u_2$. Instead of using the MIG to decide \textit{when} to switch modes, we use it to decide \textit{whether} to switch modes and allow user input. To aid in this evaluation, we consider the MIG over the entire time horizon T and thus use the integral of it as our evaluation criterion. Our approach to calculating the MIG criterion is outlined below.  

The mode insertion gradient is usually defined as 
\begin{equation}\label{mig}
    \frac{dJ}{d\lambda}(\tau)=\rho(\tau)^T \left[f(x(\tau),u_2(\tau))-f(x(\tau),u_1(\tau))\right]
\end{equation}
for a system with dynamics 
    \begin{equation*}
    \dot{x}(t) = f(x(t), u(t), t) = g(x(t)) + B(x(t), t) u(t),
    \end{equation*}
where $\dot{x}(t)$ is linearly dependent on the control $u$. In (\ref{mig}), state $x$ is calculated using nominal control, $u_1$, and $\rho$ is the adjoint variable calculated from the nominal trajectory $x(t)$,
\begin{equation*}
    \dot \rho = -\nabla l_1(x)-D_xf(x,u_1)^T\rho,
\end{equation*}
where $l_1(x,t)$ is the incremental cost and $\rho(t_0+T)=\nabla m(x(t_0+T))$ is the terminal cost. Moreover, in the work presented here, we define the nominal control, $u_1$, to be equivalent to the calculated controller action ($u_1(t) = u_{controller}$), and we define $u_2$ with the piece-wise function below,

\[ u_2(t) =  \left\{
        \begin{array}{ll}
              u_{user} & t\leq t_0+t_s \\
              u_1 & t_0+t_s <~ t\leq ~t_0+T \\
        \end{array} 
\right. \]
\\

\noindent where $t_s$ is the sampling time, $T$ is the time window over which we are evaluating system behavior, and $u_{user}$ is a user input recorded at current time $t_0$. It is worth noting that $u_2$ is defined by a combination of user input at current time $t_0$ and actions from an optimal controller over time $T$ into the future\endnote{Sequential Action Control~\citep{therakis2018} was used to compute the nominal controller action for both criteria.However, any control policy that can be computed in real-time could be used.}.
It is worth noting that $u_1$ is not a schedule of actions that is precomputed ahead of time, instead we calculate the best sequence $u_1$ every time step $t_s$ based on the previously taken action and current state of the sytem. In turn, the action sequence $u_2$ is defined by a combination of user input at current time $t_0$ and newly calculated actions from an optimal controller over time $T$ into the future.
This gives unique flexibility to the criterion and grants the user more control authority over the joint system, because any user action that could be corrected for by a future optimal action or sequence of optimal actions without destabilizing the system during the time window $T$ will be admitted. Even suboptimal user actions will be allowed.   

When using MIG as an evaluation criterion, we calculate the integral of the mode insertion gradient over a time window $T$ into the future
\begin{equation}\label{eq:migcrit}
    \int_{\textcolor{blue}{t_0}}^{\textcolor{blue}{t_0}+T}\frac{dJ}{d\lambda}(t)\delta t,
\end{equation}
to evaluate the impact of user control $u_2$ on the system over time $T$. When negative, the integral indicates that $u_2$---the user input---is a descent direction over the entire time horizon, which can be shown by evaluating the change in cost due to a control perturbation $u_2 - u_1$. Thus, the MIG integral can serve as the basis for evaluating the impact of a current user action on the evolution of a dynamic system over a time window into the future and has proven in our experiments to be a balanced assessment criterion---significantly improving performance while only minimally rejecting user actions. 

\subsubsection{Optimal Controller Inner Product Criterion.}\label{sec:IPcrit}

The optimal controller inner product (OCIP) criterion works in algorithm~\ref{algorithm} by computing the value of a nominal controller $u_c$ based on the current state of the system. In this study, we use a model predictive controller described in~\cite{ansari2016}, and when the system is near equilibrium, we switch to a linear quadratic regulator (LQR). Note that any controller could be used, but it should be capable of driving the system by itself according to the desired specification. Calculating the inner product between the user input and the nominal controller establishes whether or not the two vectors are in the same half plane (e.g. $\langle u_c, u_{user} \rangle >0 $). One can further specify that the user input vector must lie within a cone near the nominal control vector by specifying a maximum angle $\gamma$ between $u_{user}$ and $u_c$. 
If the user input lies in the same half plane as $u_c$ and within $\gamma$ radians of $u_c$, then the filter does nothing. This acceptance criterion is given by,
\begin{equation}\label{inner_product}
\langle u_c, u_{user} \rangle >0\  \textbf{and}\ |\phi|\leq\gamma.
\end{equation}
If the inner product between the control and the user command vector is positive, and the corresponding angle of the vectors is small, then the effect of user input on the system should be similar to that of the control vector.
If the user input is not in the same half plane as $u_c$ or not within $\gamma$ radians of $u_c$, the input is rejected. 

\subsection{Experimental Platform}\label{nact3d}

\begin{figure}[!t]
\centering
\includegraphics[width=\linewidth]{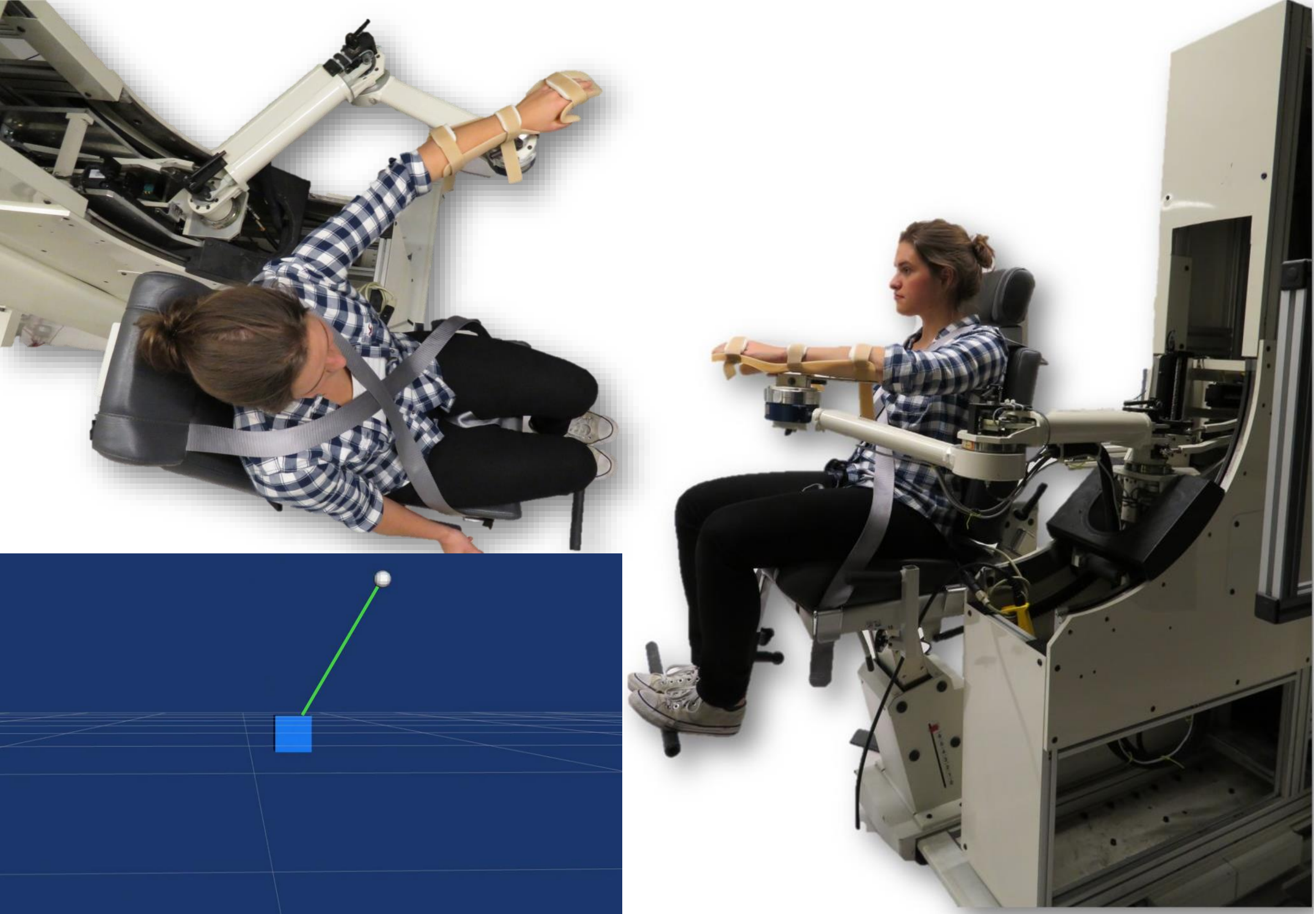}
\caption{The New Arm Coordination Training 3D (NACT-3D) device provides haptic feedback in three dimensions to simulate a specified inertial model via admittance control. A force-torque sensor at the end-effector provides input to the admittance control loop. During this experiment, high stiffness virtual springs were used to restrict user motion in the z-direction while allowing them to move freely in the x-y plane. The display (bottom left) provided real-time visual state feedback of the cart-pendulum system.}
\label{fig:nact3d}
\end{figure}

All subject data was collected using the New Arm Coordination Training device (NACT-3D) shown in Figure~\ref{fig:nact3d}. The NACT-3D is a powerful haptic admittance-controlled robot that can be used to render virtual objects, forces, or perturbation in three degrees of freedom. This device is similar to that described in \cite{stienen2011} and \cite{ellis2016}, to quantify upper limb motor impairments and  provide a means to modulate limb weight support during reaching. While in use, the subject is seated in a Biodex chair connected to the base of the NACT-3D with their arm secured in forearm-wrist-hand orthosis. The NACT-3D is capable of exerting forces  at this interaction point between the user and the robot in the x, y, and z directions only. The impedance control is updated at $1000Hz$. 

\begin{figure}[!h]
\centering
\includegraphics[width=\linewidth]{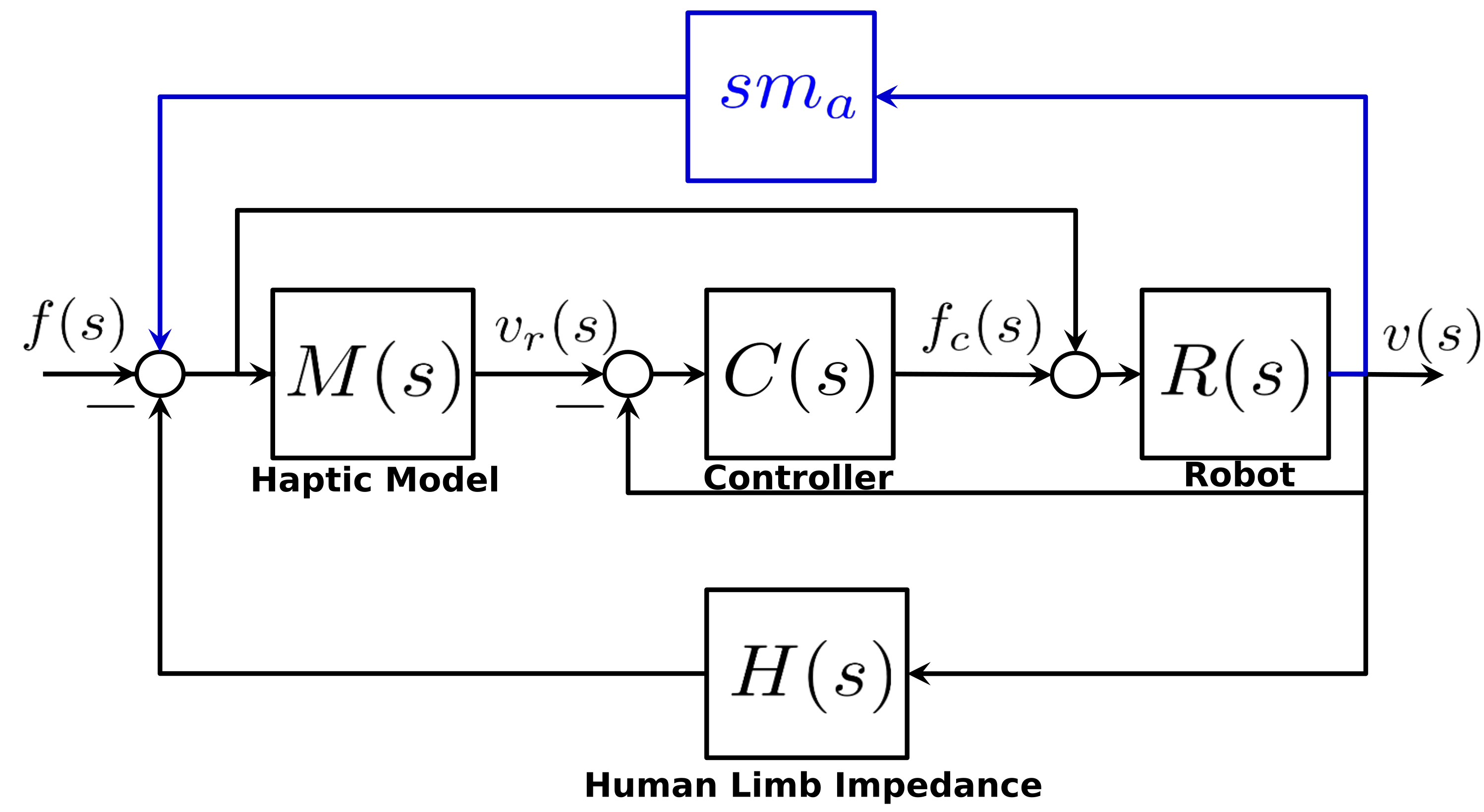}
\caption{A voluntary force $f(s)$ is measured at the robot's end-effector using a six degree of freedom force-torque sensor (JR3) and passed through a model $M(s)$ that determines the velocity $v_r(s)$ the robot should move with. The reference velocity is tracked by the low level velocity controllers of each motor drive. The human also delivers involuntary impedance forces due to movement, given by dynamics transfer $H(s)$. Acceleration information is fed back as a pseudo-force for extra inertia reduction of the system.}
\label{admittance}
\end{figure}

The NACT-3D can move its end effector within a workspace defined both by its design limits 
(a radius of approximately 0.6m around the participant’s shoulder in the half plane in front of the participant's chest) and safety limits set by the investigators. 
The splint can rotate passively but no torque can be exerted by the robot. At the point where the splint is mounted, a force-torque sensor measures the subject input which is fed back to the admittance controller. 
The peak push-pull force that can be exerted by the device of the device at the end effector is approximately $4.7kN$. The force measured at the end effector is sent to a host computer for use in the assistance algorithm to compare the user input to the control policy and perform the filter update at a rate of $60Hz$. 
In Figure~\ref{admittance}, $f(s)$ is the subject control input which is used in the filtering algorithm. At start up, the haptic model is set such that the model of the end effector accounts for the mass of the subject's arm as well as an inertia parameter defined by the investigator. 

During testing, a display provided real-time visual state feedback to the user about the cart-pendulum system s/he was trying to invert. High stiffness virtual springs in the haptic model were used to restrict user motion to a horizontal plane corresponding to the path of the cart in the virtual display. When user inputs met the criterion being used, they were accepted and the robot behaved according to the control scheme described in Figure~\ref{admittance}. When user inputs did not meet the criterion for acceptance, the user input $f(s)$ was ignored by the admittance controller, such that the robot maintained its velocity at the time of rejection. Although the device was capable of replacing the user input with an input prescribed by an optimal controller, we chose to simply reject user actions. In this way, we provide feedback only by corrections without demonstrating or guiding the user in the correct action.

\subsection{Experimental Task}

Users were tasked with controlling a simulated two-dimensional cart-pendulum system, which they were instructed to swing up to the unstable equilibrium (the system was initially resting at the downward stable equilibrium). The equations that describe the underactuated cart-pendulum system shown bottom left in Figure~\ref{fig:nact3d} are given by:
\begin{gather}
\label{eqpend}
\dot{x} = f(x,u) =  \begin{pmatrix}
  \dot{\theta} \\
  \frac{g}{l}\sin \theta + u \cos \theta - \frac{b}{m l^2} \dot{\theta} \\
  \dot{x}_c  \\
  u
 \end{pmatrix}
\end{gather}
where the state vector $x$ consists of the angular position and velocity of the pendulum and the position and lateral velocity of the cart, $x = [\theta,  \dot{\theta}, x_c, \dot{x}_c]$, the input $u$ is the lateral acceleration of the cart, $g$ is the acceleration due to gravity, $b$ is the damping coefficient, $l$ is the pendulum length and $m$ the mass at the tip. 

Users kinematically controlled the cart acceleration (and thus position) by moving their arm from left to right in the horizontal plane subject to the constraints of the admittance controller outlined in Figure~\ref{admittance}. To avoid confusion associated with conflating the task-related forces with forces generated by the assistance algorithm~\citep{powell2012}, no haptic feedback related to the system dynamics was displayed to the user during either nominal task execution or in addition to the assistance. In both the assisted and unassisted cases, users had to rely solely on visual state feedback to understand the system dynamics.
 
\subsection{Sample Response}

\begin{figure}[!h]
\centering
\includegraphics[width = \linewidth]{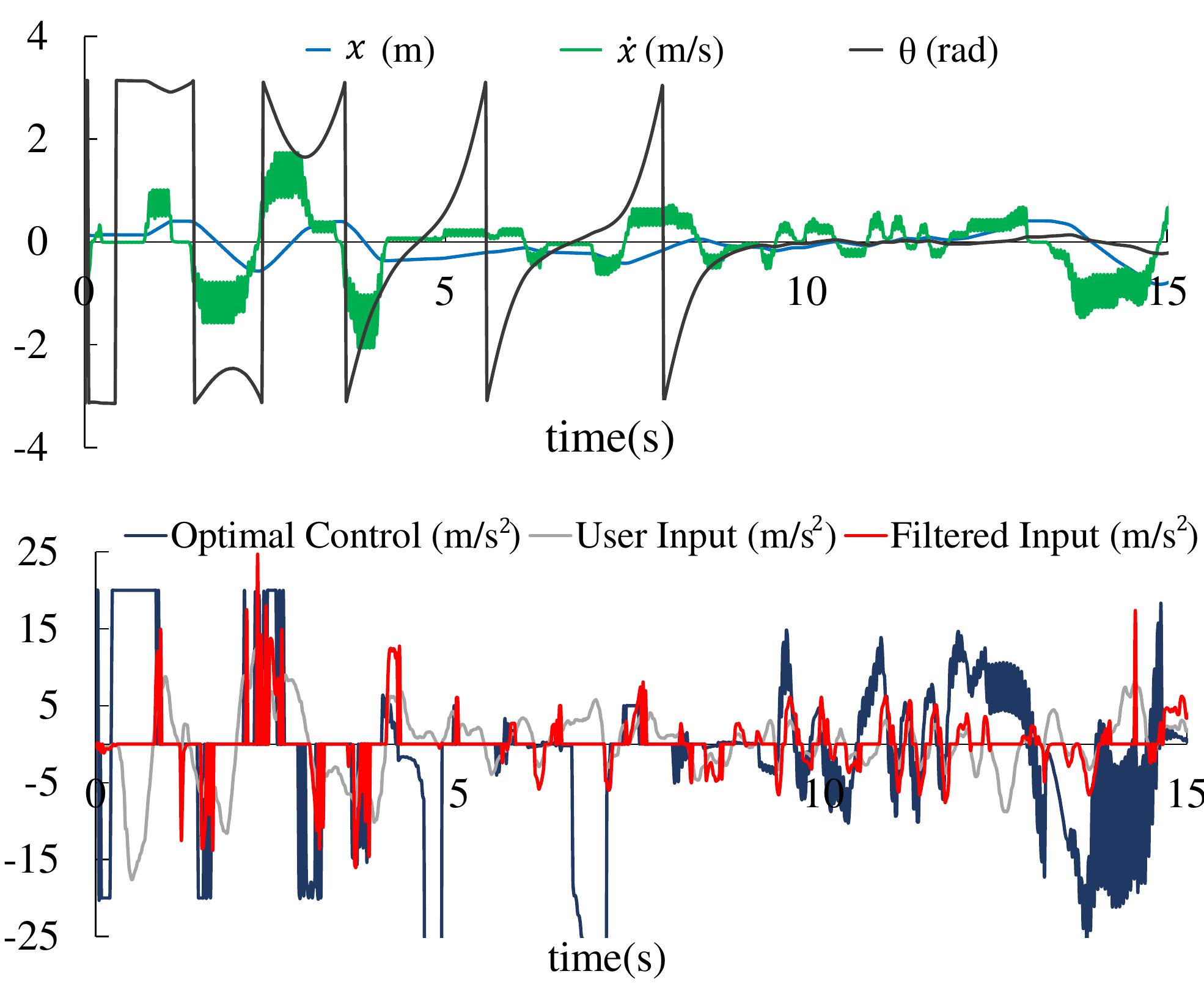}
\caption{Sample response of a subject using the NACT-3D with the OCIP criterion. The NACT-3D is able to directly shape user input. We can see that even relatively large user inputs (gray) can be reduced to zero in the filtered input (red). Top: the states of the cart-pendulum system. The subject kinematically controls the cart position $x_c$ (and $\dot{x}_c$) through the cart's lateral acceleration. We see the subject is able to stabilize the pendulum for $5s$.
Bottom: The reference signal  and user input used in \eqref{inner_product} to generate the filtered input that drives the system.}
\label{NACTsamp}
\end{figure}

The mechanical filtering imposed by the robotic platform forces changes in the user input. Figure~\ref{NACTsamp} shows a sample response of user inputs in
assistance mode. 
Shortly before $t=4s$, we see an example of a rejected user action.  Although the user input (gray) is a positive acceleration, the filtered input (red) is zero, and the velocity of the cart (green) is held constant. The optimal control signal (blue) indicated that a negative acceleration should be applied, but this was not used to replace the user input nor was it communicated to the user.
At around $t=3.5s$, the user attempts a negative acceleration, and the prescribed optimal controller is also negative. 
Under the OCIP criterion, this action is allowed and the cart velocity decreases. 
This demonstrates how the mechanical filter can effectively yield to skilled users while assisting unskilled users. 

Unlike the haptic stylus used in ~\cite{fitzsimons2016}, the robotic platform used in the studies discussed in this paper was capable of fully rejecting the physical motions of the subjects because of its underlying control architecture and sufficient actuation capabilities. 
While the haptic stylus, relied on users to interpret the feedback and correct their motion, the device described below in Section~\ref{nact3d} could actively correct motion while giving feedback and did not rely so heavily on the subjects interpretation of the haptic feedback. This allowed us to update the mechanical filter at a higher rate (60Hz-100Hz) than in the previous implementation (10Hz), which is part of why the improvements in performance are much greater on this device. In the trial shown in Figure~\ref{NACTsamp}, the user stabilizes the pendulum at the unstable equilibrium at $t=9s$ and maintains that configuration for $5s$.

\subsection{Experimental Protocol}\label{sec:exp}
Subjects used an upper limb robotic platform (NACT-3D) as an interface to control a simulated cart-pendulum system with state vector $x=[\theta, \dot{\theta}, x_c, \dot{x_c}]$ and horizontal acceleration of the cart as control input. During experimental trials, the user's goal was to invert the pendulum to its unstable equilibrium. User input was inferred from a force sensor at the robot's end-effector.
 
\begin{figure}[h]
\centering
\includegraphics[width=\linewidth]{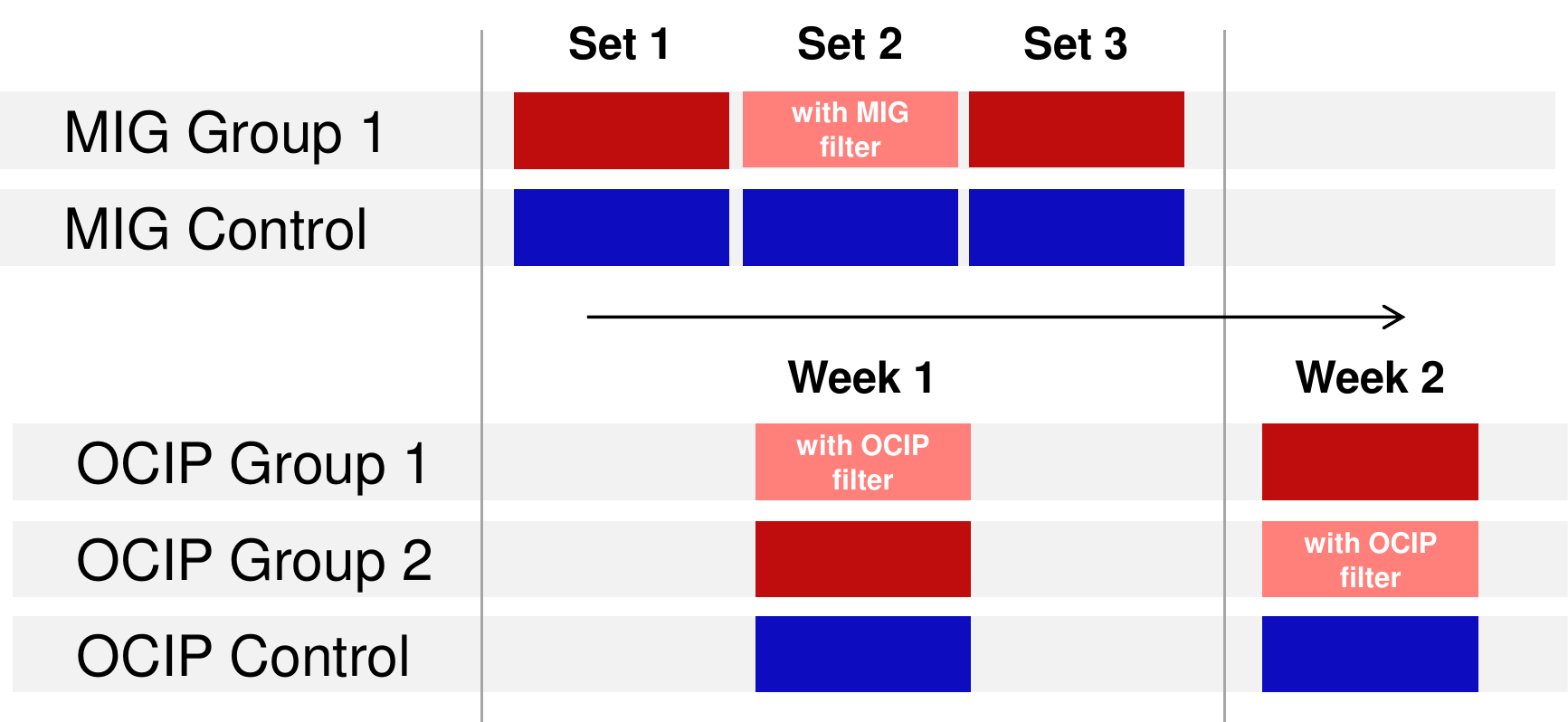}
\caption{Each rectangle represents a set of 30 trials. MIG study participants completed all sets on the same day. OCIP participants completed sets one week apart.}\label{protocol}
\end{figure}

At the beginning of each session subjects were seated and secured in a Biodex chair and their left arm was secured in the orthosis on the NACT-3D (Figure~\ref{fig:nact3d}). The system and task was demonstrated to them at the start of the testing using a video of a sample task completion. Subjects were instructed to attempt to swing up the pendulum to the upward unstable equilibrium and balance there for as long as possible. Subjects were instructed to continue to try to do this until the 30 second trial was over even if they succeeded at balancing near the equilibrium more than once. Depending on the study, subjects performed sets of 30 trials with short breaks in the same session or in sessions scheduled approximately one week apart as shown in Figure~\ref{protocol}. 

Subjects were recruited locally, and had to be healthy, able-bodied adults (in the age range of 18 to 50) with no prior history of upper limb or cognitive impairments. Only right-hand dominant participants were accepted into the study, and each subject performed the task with their left limb.
All study protocols were reviewed and approved by the Northwestern University Institutional Review Board, and all subjects gave written informed consent prior to participation in the study.

\subsubsection{MIG Study.} 
Twenty-eight subjects (9 males and 19 females) consented to participate in the MIG study. All subjects in the MIG study completed three sets of thirty 30-second trials with short breaks between sets. Upon enrollment, subjects were randomly placed into either a control ($n=10$) or training group ($n=18$). During the second set, feedback in the form of a filter  using the MIG criterion was engaged for the training group, while the control group completed each of the three sets without any feedback. Again, each user did three sets of thirty trials: set 1 (both groups: no feedback), set 2 (control: no feedback, training: feedback in the form of a mechanical filter using MIG), set 3 (both groups: no feedback).

\subsubsection{OCIP Study.}
Fifty-three subjects (17 males, 36 females) consented to participate in this study. Each subject completed 2 sessions being approximately one week apart. Upon enrollment in the study, each subject was placed into 1 of 3 groups. If placed in the training group ($n=20$), the subject completed the first session with the OCIP filter and received no assistance in the second session. If a subject was placed in the non-training group ($n=20$), they performed the task without assistance in the first session and used the OCIP filter in the second session. Finally, a control group ($n=13$) performed the task without assistance in both the first and second session.

\begin{figure*}[!h]
\centering
\includegraphics[width=0.75\linewidth]{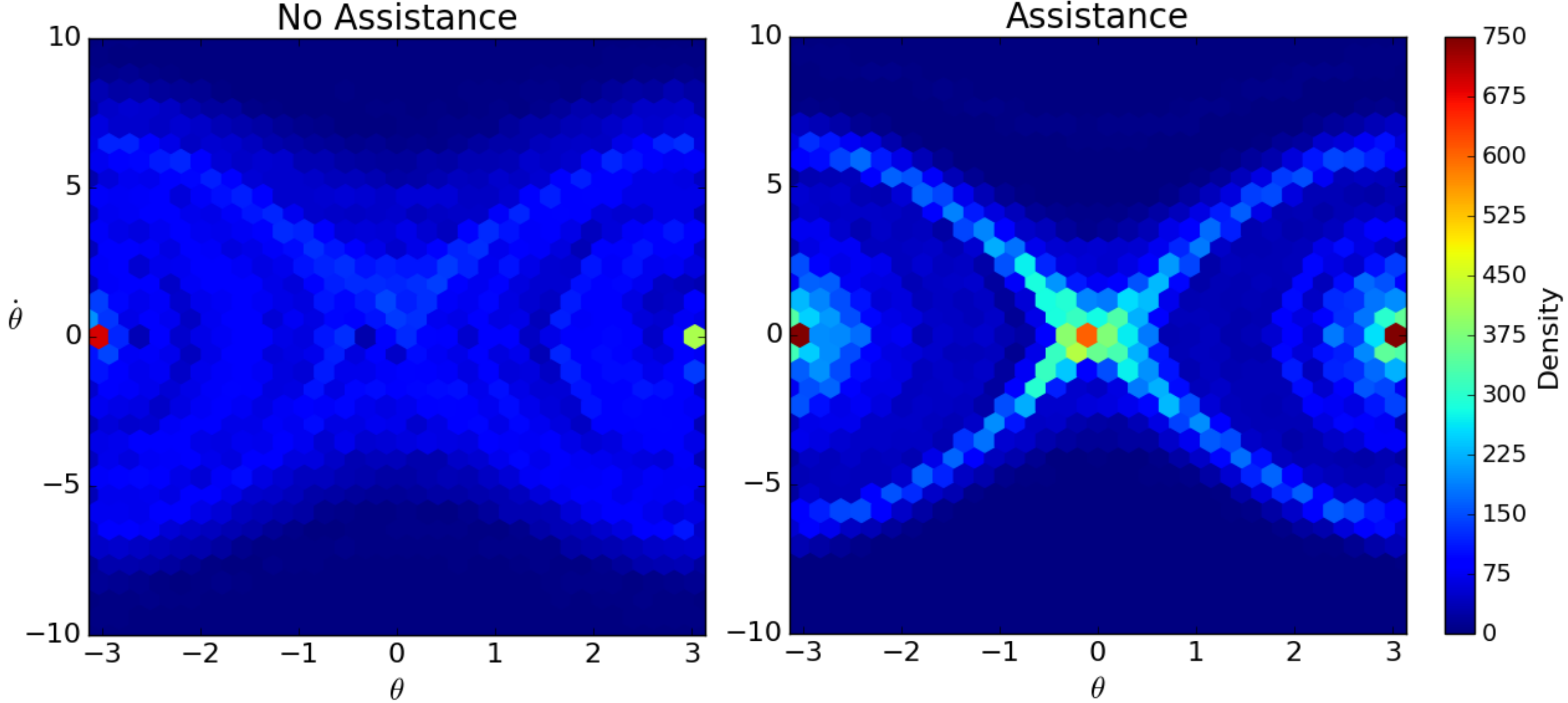}
\caption{A histogram of all the trajectories recorded in the OCIP study demonstrates how the statistics of unassisted and assisted trajectories differ from one another. The histogram of unassisted trajectories (left) has its highest density at $\theta=\pm\pi$ which is the farthest point from the goal state. The rest of the distribution is diffuse over the state space. Although the histogram of the assisted trajectories (right) also has a high density at $\theta=\pm\pi$, the distribution is not as diffuse as that of the unassisted trajectories. There are bands of high density spreading outward form the goal state $(\theta,\dot{\theta})=(0,0)$. The spatial statistics of the assisted trajectories are more similar to the reference distribution, because there is a high density at and around the goal state. This outcome is captured by measuring the distance from ergodicity of the trajectories in each group with respect to the reference distribution.}
\label{fig:assist}
\end{figure*}
\subsection{Performance Measures}\label{metrics}

The full state and user inputs were recorded in each trial and were used to calculate task-specific performance measures as well as more general measures such as error. The task-specific performance measures used to evaluate subjects in both studies is predicated on a notion of success. The definition of success that was used was based on the region of attraction for a linear quadratic regulator capable of stabilizing the system dynamics defined in the experiments.
A trial was considered successful when a subject reached an angle of 
$\pm0.15$~rad and angular velocity of $\pm0.6$~rad/s. 
This definition of success was used to determine the time to success of the users in each experiment. In addition, if a subject was successful, the total time spent at the angle and angular velocity defined as success was recorded as the balance time. When users were successful multiple times in the same trial, time spent in the balance region was cumulative. 

While these outcome-based measures provide clear indication about whether or not users could meet task goals, they neglect the behavior of users away from the goal state. Therefore, we use two measures---error and ergodicity---that use the full trajectory data to characterize task performance. The root mean square (RMS) error of each trajectory generated by the users was calculated with respect to the desired position in an inverted unstable equilibrium (zero-vector of the states). RMS error was normalized by the RMS error of a constant trajectory at the stable equilibrium, equivalent to the error of the user not moving from the initial conditions. Finally, we also compared the experimental conditions through an analysis of the spatial distribution of trajectories that we observe under each condition. For instance, in the histogram  of states recorded for all subject trajectories (Figure~\ref{fig:assist}), one can see that trajectories in which subjects received assistance have high density values near the goal state. 
To quantify the comparison of the distributions, we compute a metric on the ergodicity~\citep{mathew2011,miller2016} of each trajectory with respect to a Dirac delta $\delta(x-s)$ function centered at the unstable equilibrium $(\theta,\dot{\theta})=(0,0)$. The ergodic measure captures how well the time averaged statistics of the trajectory match the statistics of the reference distribution. The value of this metric was determined by calculating the weighted distance between the Fourier coefficients of the trajectory and those of the distribution. The ergodic metric gives us the distance from ergodicity, such that trajectories which were highly ergodic had lower ergodicity than those that were less ergodic. 

The controller intervention was measured as the percent of rejected actions (PRA). PRA measured the fraction of user inputs that were rejected, where we defined an action to be a non-zero user input.

\subsection{Statistical Analysis}
The MIG experiment consisted of 30 baseline trials, 30 trials with or without the MIG filter, and 30 trials post-training for a total of 90 trials. These were grouped into blocks of 5 trials to evaluate subject performance over time. The analysis consisted of two-factor (block and group) repeated measures ANOVA tests, using the baseline and post-training data only. The ANOVA's were used to compare the effect of the MIG filter and unassisted practice on each of the performance measures. Trials from set 2 are removed from the analysis to avoid including the assistance itself as a factor in the experiment. 
In the OCIP study, subjects trained with the filter received no prior exposure to the task without assistance. Student's t-tests were used to evaluate the difference between the week 2 performance of the trained group and the week 2 performance of the control group. 

\begin{figure*}[!h]
	\begin{center}
		\includegraphics[width=\textwidth]{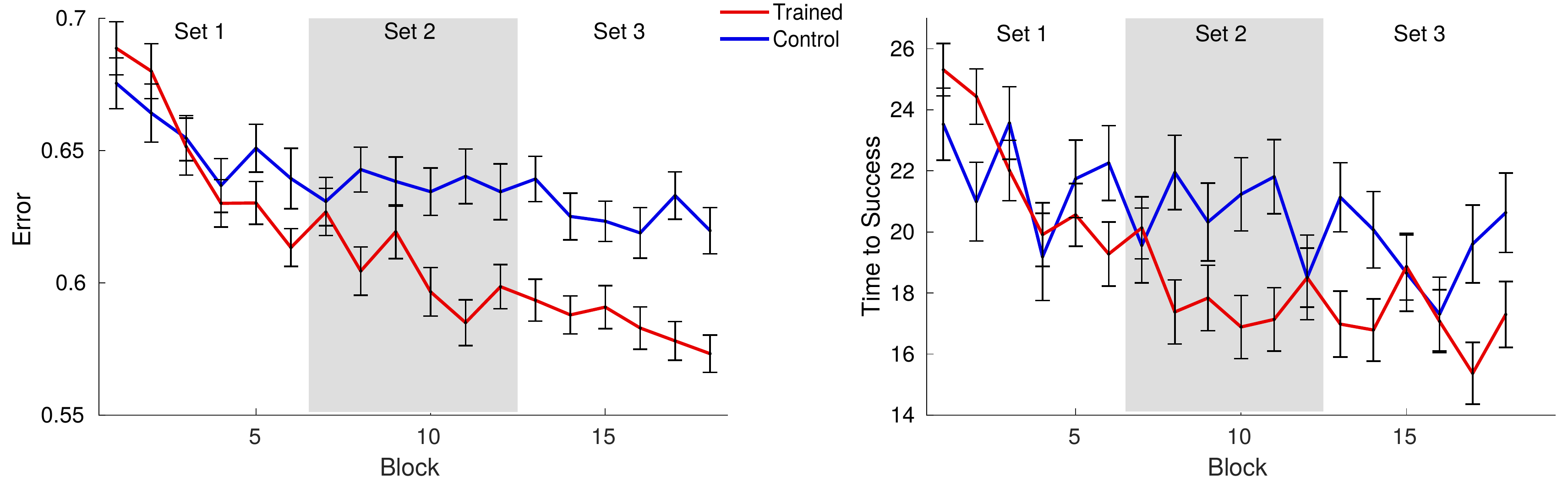}
	\end{center}
	\caption{ The MIG study showed that subjects improved with practice in all sets regardless of training group, however, there was a significant interaction effect between training group and block when ANOVAs were applied to three of the four performance metrics. This suggests that although subjects in each group started around the same performance level, the trained group attained a higher level of performance than the the control group during the post-training trials. Note that the set 2 performance (gray) was not included in the ANOVA to avoid measuring effects of the assistance itself.}
	\label{fig:MIGanova}
\end{figure*}

The relevant features of the hybrid shared controller were evaluated statistically. First, the ability of the shared controller to assist subjects in completing the task was tested in each study. In the MIG study, this was done by comparing the experimental group to controls with an equivalent amount of practice using a two-sample t-test. The effect of the OCIP criterion as an assistive controller was tested in a counter-balanced fashion using paired two-sample t-tests on all performance metrics. Second, the sensitivity of the shared controller to the initial skill of the users was evaluated by performing Peason's R correlation tests between the level of controller  intervention and the performance of users in their first set of unassisted trials. Finally, the assist-as-needed feature of the shared controller was shown by testing the correlation between the level of controller intervention and the current performance of subjects.

\section{Results}\label{sec:results}

The results were reported as follows. First, the training effect of each study was statistically tested in Section~\ref{sec:train}. The results demonstrated that training with the hyrbid shared controller increased subject performance in later trials within the same session (Section~\ref{sec:MIGtrain}) and in a session one week after training (Section~\ref{sec:OCIPtrain}). An analysis of the hyrbid shared controller was performed to test for three characteristics of effective pHRI. In Section~\ref{sec:assist}, the performance improvement made while the criterion was engaged was evaluated in both the MIG study and the OCIP study. In Section~\ref{sec:initskill} and \ref{sec:aan}, the correlation of the percent of rejected actions with the initial skill and current performance are reported to evaluate the sensitivity of the shared controller to user skill and its ability to assist-as-needed, respectively. In each section the relevant statistics are reported first, followed by a summary and interpretation of the results.

\subsection{Training Effect}\label{sec:train}
The effectiveness of the filter as a training tool was assessed in both experiments. In the MIG study, we consider only skill acquisition within a single session. We assess the retention of skill over the course of one week in the OCIP study.
\subsubsection{MIG Study: Skill Acquisition.}\label{sec:MIGtrain}

Two-factor repeated measures ANOVAs were used to assess the effects of the group (between-subjects) and set (within-subjects) on all performance measures listed in Section~\ref{metrics}. The training group and control group were evaluated based on the baseline trials (set 1) and the post-training trials (set 3) only. Set 2 was left out of the ANOVA, so that effects of the assistance itself would not be measured in the analysis. In order to assess how subject performance evolved over time, the baseline and post-training sets were analyzed using blocks containing five individual trials. Therefore, there were 6 blocks in each set as shown in Figure~\ref{fig:MIGanova}.  

The factorial ANOVA of the balance time revealed that block was the only significant factor ($p<2\times 10^{-16},\ F(11,286)=10.775$). The main effect of group and interaction effect of group and block were not significant for balance time ($p>0.05$). When an analysis of variance was performed on the time to success, again, the main effect of block was significant ($p=3.81\times10^{-15},\ F(11,286)=9.848$) and the main effect of group was not significant ($p=0.533,\ F(1,25) = 0.399$). However, the interaction effect of group and block was significant ($p=0.0135,\ F(11,286) = 2.222$). The control and trained group performed similarly in the baseline trials. The time to success decreased even before the training set (Figure~\ref{fig:MIGanova}). However, the control group essentially plateaued during the training set and saw large fluctuations in the time to success during the post-training trials. The time to success of the trained group decreased during training and was maintained in the post-training trials.

The group also was not a significant factor affecting the RMS error ($p=0.223,\ F(1,25)=1.560$), but main effect of subset ($p<2\times10^{-16},\ F(11,286)=20.620$) and the interaction of group and subset ($p=0.004,\ F(11,286) = 2.575$) were significant. When the error of the control group and trained group was plotted over time (Figure~\ref{fig:MIGanova}), the control group error decreased initally but leveled off. The error of the trained group continued to decrease during training and in the post-training trials. 

When the distributions of the trajectories were compared using the ergodic metric, the significant factors were the subset ($p<2\times 10^{-16},\ F(11,286)=18.311$) and the interaction between group and subset ($p=0.030,\ F(11,286) = 1.983$). The main effect of group was not significant ($p=0.294,\ F(1,25)=1.151$). The progress of the ergodic metric over time was similar to that of the RMS error.
 
\begin{figure*}[!h]\centering
\begin{subfigure}{0.45\textwidth}
\centering \captionsetup{width=0.9\linewidth}
\includegraphics[width=\linewidth]{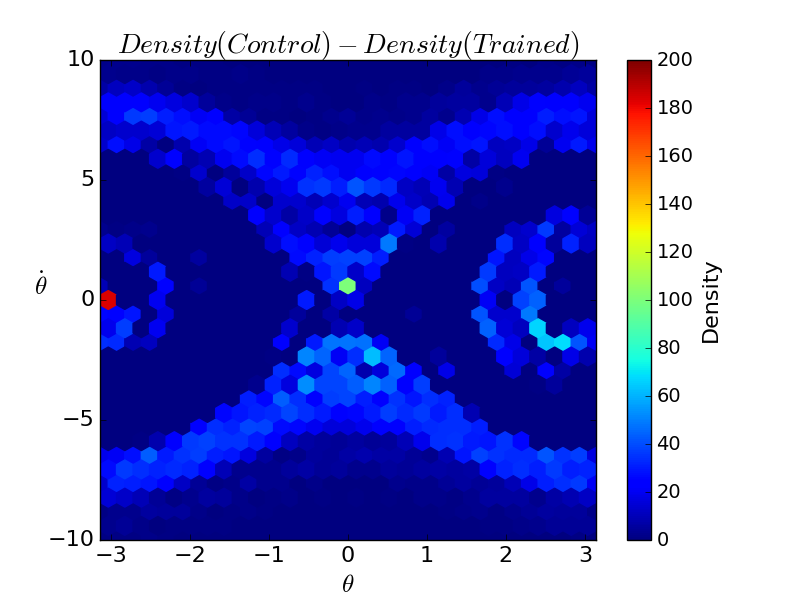}
\caption{The density function of trained group trajectories subtracted from the control trajectories density.}
\end{subfigure}
\begin{subfigure}{0.45\textwidth}
\centering \captionsetup{width=0.9\linewidth}
\includegraphics[width=\linewidth]{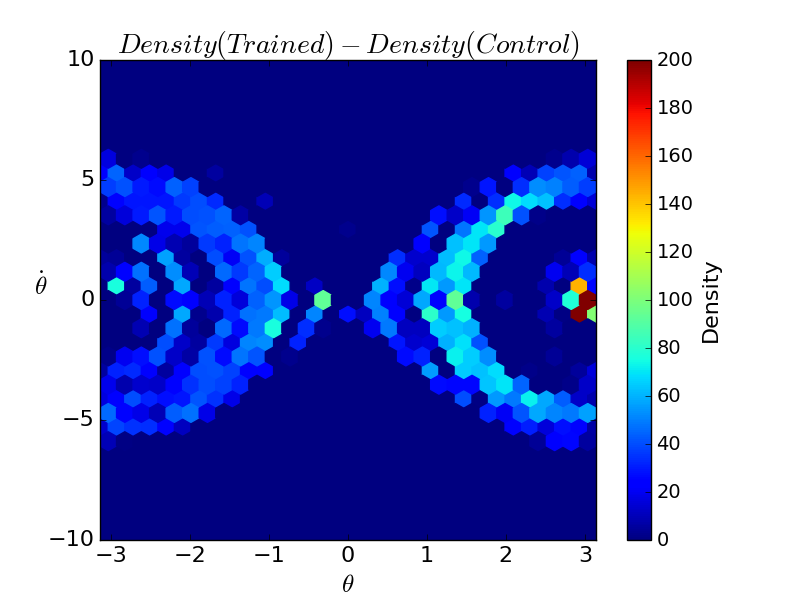}
\caption{Control trajectories density function subtracted from the post-training trajectories density.}
\end{subfigure}
\caption{Trajectories from Week 2 of the OCIP study showed that subjects who trained with the hyrbid shared controller spent more time near the goals state $(\theta,\dot{\theta})=(0,0)$ than subjects who practiced unassisted. On the left, the week 2 control trajectories have higher densities than the post-training trajectories at higher angular velocities as well as in bands near $\theta=\pm\pi$ which is the farthest angle from the goal state. The control trajectories also spend time near the goal state, but to a lesser extent. On the right, the trained trajectories also have high density near $\theta=\pm\pi$, but there are large bands of high density in the region $-1.5\leq\theta\leq 1.5$ and $-4\leq\dot{\theta}\leq 4$. 
This suggests that the trained group's motions were more consistent with the task goal---making the statistics of the trained group closer to the spatial statistics of the reference Dirac delta distribution, so the ergodic measure of the trained group is lower than that of the controls.}
\label{fig:learn}
\end{figure*}
\emph{The results of the ANOVA of each of the performance measures showed that subset was a significant factor---implying that regardless of the training in set 2, all subjects performed better in later sets than in their initial sets. The significant interaction effect observed in three out of the four metrics demonstrates that while the subjects started at the same performance level, subjects in the trained group attained a higher performance level than the control group.} 

\subsubsection{OCIP Study: Short-term Retention.}\label{sec:OCIPtrain}

The effect of training was assessed by comparing the week 2 session of the trained group to the week 2 session of the control group. The two groups were not significantly different in terms of the task-specific measures of success. However, the trained group had significantly lower RMS error, and the distributions of the trained group's trajectories were more similar to the reference distribution, resulting in a much lower ergodic measure than the control group.  A two-sample t-test was performed on the task specific performance measures, finding no difference between trained group and untrained group in terms of their time spent balanced ($p=0.1687,t(988)=1.378$) and time to success ($p=0.1935,t(988)=1.301$). The two-sample t-test of the RMS error showed a significant difference between the trained ($mean = 0.621 ,\ SD=0.058$) and control ($mean=0.629,\ SD=0.061$) groups ($p=0.0499,t(988)=-1.963$). The t-test of the ergodic metric also showed a significance difference ($p=2.266\times10^{-4},t(988)=-3.701$) between the trained group ($mean=0.705,\ SD=0.177$) and the control group ($mean=0.751,\ SD=0.207$). Although subjects who trained with the OCIP criterion were not successful more often than the control group, they did spend a higher proportion of their time near the goal state as can be seen by the histogram of their trajectories shown in Figure~\ref{fig:learn}. \emph{These results suggest that subjects learned more and retained that skill one week after training when they trained with assistance rather than simply practicing the task unassisted.}

\begin{figure*}[h]
	\begin{center}
		\includegraphics[width=\textwidth]{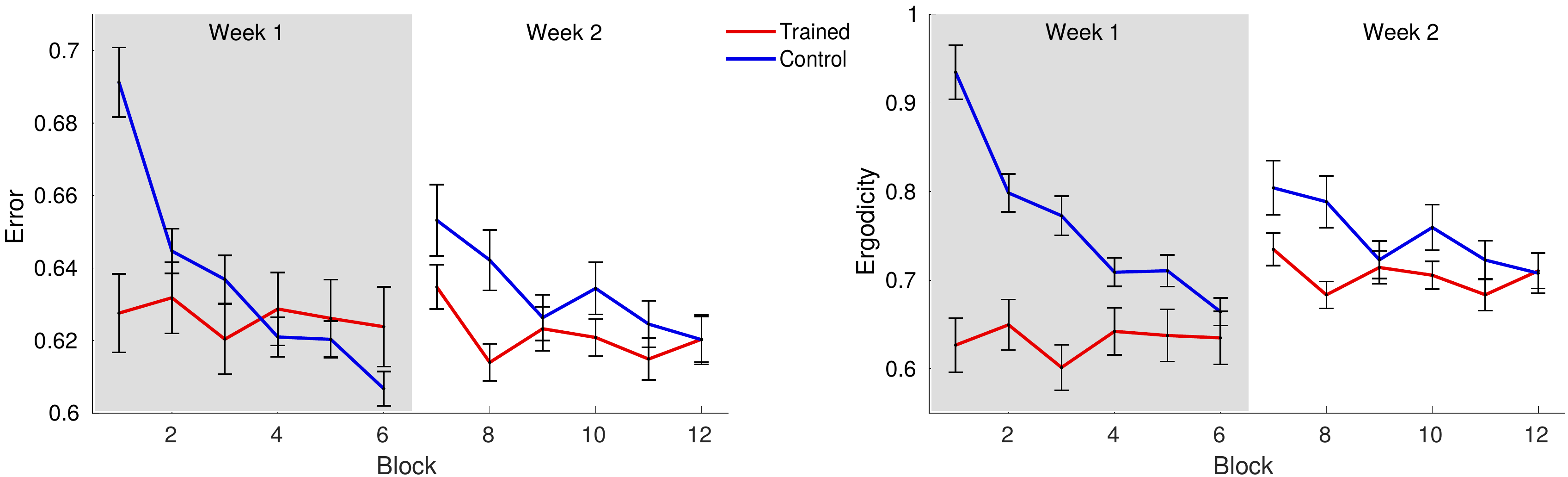}
	\end{center}
	\caption{ The results of the OCIP study demonstrate that subjects trained in week 1 retain high performance levels in week 2 as measured by RMS error and ergodicity. In the first 2 blocks of trials, the error and ergodicity of the control group are higher than that of the trained group. The trained group retains their initial performance level, while the control group continues to improve---eventually reaching the same level of performance as the trained group. It appears the feedback helped with retention because the learning was more structured. Note that the performance measures in week 1 (gray) were not used in the statistical analysis to avoid measuring the effects of the assistance itself.}
	\label{fig:OCIPanova}
\end{figure*}

The progress of the two groups over the second session (Figure~\ref{fig:OCIPanova}) was analyzed further by performing mixed design ANOVAs on the training group (between participants) and block (within participants) using all four measures. 
 
 The balance time of the control group and the trained group in the second session was analyzed with a 2 (training groups) x 6 (blocks) mixed design ANOVA, which showed no significant main effects or interactions effects. The main effect of training group was not significant $F(1,31)=1.202,\ MSE = 1.25,\ p=0.28,\ Cohen's f=0.08$. The main effect of block also was not significant $F(5,155)=2.018,\ MSE = 0.44,\ p=0.079,\ Cohen's f=0.11$, nor was the interaction of training and block significant $F(5,155)=1.05,\ MSE = 0.23,\ p=0.39,\ Cohen's f=0.08$.
 
 The mixed design 2 x 6 ANOVA design was also applied to the time to success, and the main effect of training group was not significant $F(1,31)=0.334,\ MSE = 103.4,\ p=0.567,\ Cohen's f=0.05$. The main effect of block was not significant either $F(5,155)=1.34,\ MSE = 66.32,\ p=0.25,\ Cohen's f=0.09$. The interaction effect of block and training group also was not significant $F(5,155)=1.34,\ MSE = 66.50,\ p=0.25,\ Cohen's f=0.09$.
 
The same mixed design ANOVA was used to analyze the RMS error in each trial. The main effect of block was significant $F(5,155)=4.336,\ MSE = 0.011,\ p=0.001,\ Cohen's f=0.19$, but the main effect of training was not significant $F(1,31)=0.76,\ MSE = 0.035,\ p=0.39,\ Cohen's f=0.15$. The interaction effect of training group and block also was not significant $F(5,155)=1.61,\ MSE = 0.004,\ p=0.16,\ Cohen's f=0.12$.

The analysis of the ergodic metric using the mixed design ANOVA revealed a significant main effect of block $F(5,155)=2.88,\ MSE = 0.08,\ p=0.0163,\ Cohen's f=0.15$, and a significant interaction effect of block and training group $F(5,155)=2.33,\ MSE = 0.06,\ p=0.045,\ Cohen's f=0.14$. The main effect of training was not significant $F(1,31)=1.056,\ MSE = 0.49,\ p=0.312,\ Cohen's f=0.17$.

In Figure~\ref{fig:OCIPanova}, the control group performed worse at the beginning of the second session that it did at the end of the first session, and their performance increased in terms of error over the course of the session. The trained group also improves moderately during the second session. The ANOVA of the ergodic metric is also able to detect the significant improvement during the second session by the control group as well as the interaction effect of group and training. This interaction is a result of the trained group performing better under the ergodic metric at the beginning of the second session and maintaining that performance, while the control group eventually reached the same level of performance. \emph{Training with the OCIP criterion in week 1 speeds learning, and skill is retained after one week though the improvements due to unassisted practice are not retained.}

\subsection{Task-based Assistance}\label{sec:assist}
We evaluate the ability of the hybrid shared controller to assist subjects in completing the task while it is engaged. In the MIG study, we compare the the control group to the group recieving assistance during their second set of trials. In the OCIP study, the order in which subjects received assistance was counterbalanced, such that subject performance in the assisted session was compared to the same subject's performance in the unassisted session.
\subsubsection{MIG Study.}
\begin{figure}[b]
	\begin{center}
		\includegraphics[width=\columnwidth]{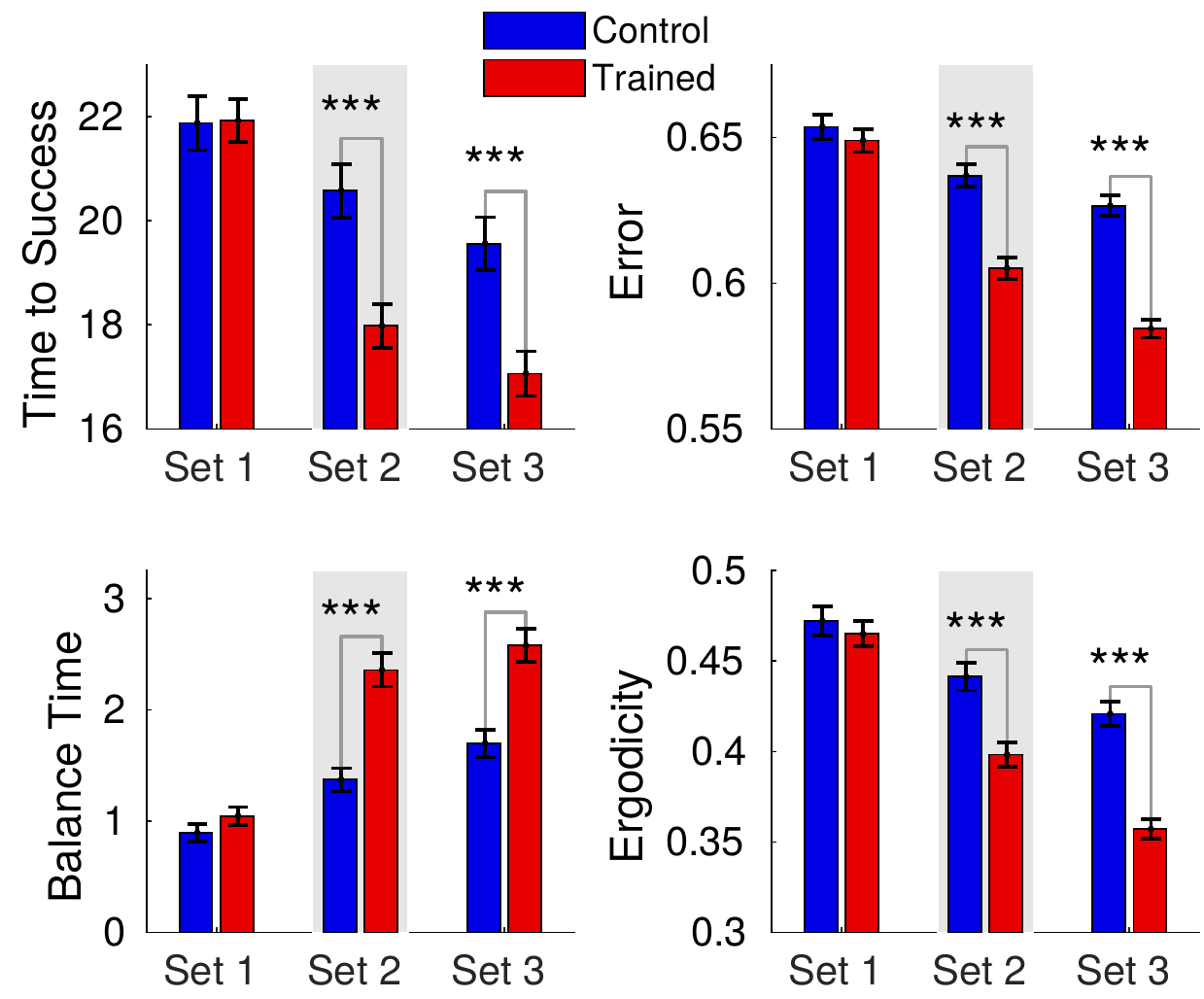}
	\end{center}
	\caption{ The MIG filter study demonstrated that the filter successfully assisted subjects in set 2 compared to controls. Moreover, trained subjects outperformed the control group in set 3. Note: error bars indicate standard error; significance is indicated by $^*p<0.05$, $^{**}p<0.01$, $^{***}p<0.001$.}
	\label{fig: MIGbar}
\end{figure}
Comparisons between the control and experimental groups are shown in Figure~\ref{fig: MIGbar}.
Two-sample t-tests showed that there was no significant difference between the control group ($n=10$) and experimental group ($n=18$) baseline performance in terms of their balance time ($p=0.178,\ t(793.22)=1.35$), time to success ($p=0.9497,\ t(644.23)=0.063$), error ($p=0.411,\ t=749.28 = -0.822$), or ergodicity ($p=0.507,\ t(711.17)=-0.6631$). 
During the training set (set 2), the experimental group ($mean = 2.36,\ SD=3.47$) maintained the pendulum in the balanced position for significantly longer ($p=7.674\times 10^{-8},\ t(832.55)=5.42$) than the control group ($mean=1.37,\ SD=1.78$). The group receiving assistance also reached the balance position more quickly than the group practicing the task without assistance ($p=9.87\times 10^{-5},\ t(666.93) = -3.9174$), so the experimental group ($mean = 17.98,\ SD = 9.79$) had a lower time to success than the control group ($mean = 20.58,\ SD=8.84$).
\begin{table*}[!h]
\centering
\begin{tabular}{lrrrrrrrr}
 &\multicolumn{2}{c}{No Assistance} &\multicolumn{2}{c}{Assistance}\\
 &\multicolumn{4}{c}{$n=40$} \\
Measure & $\mu$ & SD & $\mu$ & SD & $\mathbf{\Delta\mu}$ & t & df & $p$\\
Success Rate & $0.348$ & $0.163$  &$0.792$ & $0.182$ &$\mathbf{0.444}^{***}$ &$12.314$ & $39$ & $5.162\times 10^{-15}$\\
Balance Time & $0.191$ & $0.411$  &$1.661$ & $1.913$ & $\mathbf{1.47}^{***}$ &$26.519$ & $1199$ & $2.541\times 10^{-122}$\\
Time to Success & $25.333$ & $7.648$  &$20.068$ & $7.824$  &$\mathbf{-5.265}^{***}$ &$-17.202$ & $1199$ & $1.926\times 10^{-59}$\\
Error & $0.632$ & $0.062$ & $0.626$ & $0.102$ & $\mathbf{-0.006\ \ \ \ \ }$&$-1.674$ & $1199$ & $9.477\times 10^{-2}$\\
Ergodicity  & $0.739$ & $0.191$ &$0.631$ & $0.283$ &$\mathbf{-0.108}^{***}$ &$-11.261$ & $1199$ & $4.954\times 10^{-28}$
\end{tabular}
\caption{The OCIP filter assisted subjects in completing the task more frequently and at a higher level of performance in four out of five measures when subjects were randomly assigned to use the filter in either the first of second session. Paired two-sample t-tests were performed in \textit{R}~\citep{R} comparing the unassisted and assisted trials of the 20 subjects receiving subjects in the first session and the 20 subjects receiving assistance in the second session. Significant differences in means are indicated by $^*p<0.05$, $^{**}p<0.01$, $^{***}p<0.001$. Note that the degree of freedom (df) for success rate is 39 since there is only one rate per subject. }
\label{tab:assist}
\end{table*}
The RMS error of the experimental group ($mean = 0.605,\  SD=0.087 $) was also significantly lower ($t(753.59)=-5.925,\ p=4.738\times10^{-9} $) than that of the control group ($mean = 0.636,\ SD= 0.066$). 
Finally a comparison of the trajectory distributions of the experimental group in terms of ergodicity ($mean=0.398 ,\ SD=0.157$) to the distributions of the control group ($mean = 0.441,\ SD=0.131$) showed that the filter was effective able to effectively assist subjects in the task ($t(707.63)=-4.2435\ p=2.494\times10^{-5}$). 

The two experimental groups performed similarly in their baseline trials, but in set 2, the group using the filter outperformed the control group in terms of balance time, time to success, RMS error, and the ergodic metric.
\emph{This demonstrates that the hyrbid shared controller using the MIG criterion meets the basic requirement of assisting subjects with the task while in use.} 

\subsubsection{OCIP Study.}

In the study of the OCIP criterion, subjects were randomly placed into either a group who used the shared controller in the 1st session ($n=20$) or a group who used the shared controller in the second session ($n=20$). Therefore, the ability of the hybrid shared controller to provide assistance was tested in a counterbalanced fashion. Pairwise student's t-test were used to compare performance with and without the assistance of the filter on a subject by subject basis. Subjects did not have significantly lower error ($t(1199)=-1.674,\ p=0.0949$) when using the OCIP filter ($mean=0.626,\ SD=0.102$) compared to unassisted trials ($mean=0.632,\ SD=0.062$). Under all other metrics, subjects performed better on the day that they used the OCIP filter compared to their performance on the day they performed the task without assistance ($p<10^{-14}$) as shown in Table~\ref{tab:assist}. \emph{These results showed that the shared controller with the OCIP criterion was able to help subjects complete the task more frequently.}

\subsection{Hybrid Shared Control Adapts to Initial Skill}\label{sec:initskill}
We previously reported  that there was a relationship between participant skill level---estimated based on performance in unassisted trials---and the frequency of controller intervention in the MIG filter mode in \cite{kalinowska2018}. In that case, we calculated the success rate of the 30 trials from set 1 to approximate user skill level. We then used \textit{Percent of Rejected Actions} (PRA) values from individual trials in set 2 from the same users to identify the correlation. 
\begin{table}[h]
\begin{tabular}{lrrc}
\textbf{Measure}& Test Sign & $\mathbf{r}$ & $p$ \\
Success Rate& $-$ & $\mathbf{-0.235}$ &  $5.898\times 10^{-9}$ \\
Balance Time & $-$& $\mathbf{-0.427}$ &  $<2.2\times 10^{-16}$ \\
Time to Success & $+$& $\mathbf{0.2444}$ & $1.2898\times 10^{-9}$ \\
Error & $+$& $\mathbf{0.302}$ & $4.308\times 10^{-14}$\\
Ergodicity & $+$ & $\mathbf{0.282}$ & $2.078\times 10^{-12}$
\end{tabular}
\caption{There were moderate correlations between the initial skill of the user and PRA of the OCIP filter in all measures ($p<0.05$). Pearson's correlation tests were performed in \textit{R}~\citep{R} by applying a linear model to the mean of performance metrics first session  and percent of action rejected by the OCIP filter. The expected sign of the correlation coefficient ($r$) for a shared control scheme that is sensitive to the initial skill of the user is indicated in the column on the right.}
\label{tab:initskill}
\end{table}

The PRA of OCIP filter had a moderate correlation to the initial skill of the subjects under all of our performance measures. We evaluated the correlation between initial skill of the untrained group ($n=20$) who received no assistance in week 1 and the PRA in individual trials  of the group when they did not receive assistance from the filter in week 2. We found that there is again a significant correlation between the initial skill of the users as measured by the success rate and mean performance measures in week 1 and the PRA of those subjects in week 2. In this case, the correlation coefficients, shown in Table~\ref{tab:initskill}, were slightly higher, indicating a moderate correlation between the subject's initial performance and the filter's response to their inputs. The correlations of each performance metric matched the expected sign corresponding to a decrease in PRA in response to an increase in the user's initial skill. \emph{Although the hybrid shared controller is not tailored to either high skill or low skill users, it adapts to user skill level and could be appropriate for both novices and expert users.}

\subsection{Hybrid shared control Assists-As-Needed}\label{sec:aan}
In addition to testing the relationship between the initial skill of the user and the level of controller intervention, the responsiveness of the controller to user performance in the current trial was tested using Pearson's product-moment correlation. There were high significant correlations between user performance within a single trial and the PRA in that trial. These correlations and significance values are reported in Table~\ref{tab:currperf}. The test sign indicated in the table indicates the expected sign of the correlation coefficient when the controller accepts more user inputs in response to high user performance. Under each metric, the correlation meets this expectation. \emph{This demonstrates that the robotic assistance adapts in real-time to the needs of the users without including high-level performance heuristics to tune the relative contributions of the human and the robot.}
\begin{table}[h]
\begin{tabular}{lrrc}
\textbf{Measure} & Test Sign & $\mathbf{r}$ & $p$ \\
Balance Time & $-$& $\mathbf{-0.616}$ &  $<2.2\times 10^{-16}$ \\
Time to Success & $+$& $\mathbf{0.602}$ & $<2.2\times 10^{-16}$ \\
Error & $+$ & $\mathbf{0.677}$ & $<2.2\times 10^{-16}$\\
Ergodicity & $+$ & $\mathbf{0.706}$ & $<2.2\times 10^{-16}$
\end{tabular}
\caption{The PRA of the OCIP filter was highly correlated with the current performance of the users under all measures ($p<0.05$). Pearson's correlation tests were performed by applying a linear model to the performance measures in each trial in the OCIP study and the PRA in the same trials. The expected sign of the correlation coefficient ($r$) for a shared control scheme that is sensitive to the performance of the user is indicated in the column on the right.  }
\label{tab:currperf}
\end{table}

\section{Discussion}\label{sec:disc}
Despite the breadth of research, there are relatively few instances where physical human robot interaction has been significantly more effective than unassisted practice or human-mediated training. In the work presented here, experimental results demonstrate that our implementation of a task-based hybrid shared control paradigm enhances the effect of training compared to unassisted practice. On average, subjects who trained with our robotic feedback improved significantly more than subjects who trained with an equivalent amount of unassisted practice. Based on analysis of the spatial statistics of the post-training trajectories, the training group was capable of more controlled movement with significantly more time spent near the goal state. 
Moreover, subjects who trained with the proposed MIG shared control scheme continued to improve even after the assistance was removed, while members of the control group plateaued in their performance. Finally, through our studies, we observed that subjects both experienced immediate improvement from training with feedback and exhibited short-term retention of the acquired skill. These results demonstrate that the proposed hybrid shared control paradigm enhances task learning through forceful interaction. 

In order to understand why the algorithm was effective, we examine the unique characteristics of the hybrid shared control paradigm as well as qualities that coincide with existing best practices in robotic training.
Reviewing the motor learning literature, several features of pHRI can be identified to lead to effective training. For one, a necessary condition for effective training through forceful interaction is that the automation should be able to assist subjects in completing a task while assistance is engaged. In our experimental results, we show that the hybrid shared control paradigm  is capable of improving success in accomplishing a dynamic task during the trials in which it was engaged. In the MIG study in set 2, subjects performed better across all metrics when assistance was engaged, even though on average they started off at the same skill level in set 1. Similarly, the subjects in the OCIP study performed better with assistance compared to their own unassisted trials.

Secondly, interfaces should avoid user passivity and require substantial user effort. This is inherent to our algorithm because the hybrid controller never actively assists with task completion by only rejecting, but not replacing, incorrect actions. As a result, users are allowed to fail at the task and when they succeed, they succeed through their own actions.  While impedance-based assist-as-needed controllers can interfere less based on performance heuristics, impedance control is based on desired velocity profiles rather than the task goal. 
The hybrid shared control paradigm discussed in this paper uses a task-based criterion in order to measure whether or not it is needed. 
This allows the controller to effectively get out of the way when users are progressing towards the task goal on their own---maximizing their effort.

Building on the principle of requiring effort from the patient, shared control paradigms have been shown to be more effective when they adapt the level of assistance over time, assisting only as much as is necessary. The need for modulating the level of assistance can be due to two factors: (1) differing initial user skill level and (2) varying user performance over time. Users are expected to progress in their training over time. However, it is not enough for the level of assistance to decrease over time or after a certain performance target had been reached---there are cases, where subjects fatigue or become less engaged if the task is too difficult, so interfaces must be able to adjust both up and down in response to the automation's current assessment of the user. In our results, we show that the proposed shared control paradigm adapts to user initial skill and exhibits properties of an `assist-as-needed' controller, reducing or increasing its intervention according to user performance in real-time. In future studies, it would be interesting to explicitly assess fatigue in between or during trials. In this way, we could adjust assistance based on current levels of fatigue and/or control for the effects of fatigue in study outcomes.

All in all, we present here a hybrid shared control paradigm that significantly improves task learning. We use a task-based criterion to discretely switch between full user control and full rejection of user control, which allows us to synthesize an interface with characteristics important for motor learning. Experimental data confirms that the shared control scheme exhibits these characteristics. 

We also found that within a single session, trained subjects attained a higher level of performance than their counterparts who practiced unassisted. Yet at the end of the second session in week 2, control subjects reached the same level of performance as the trained group. This is likely due to the difference in when the hybrid shared control was introduced, and indicates an opportunity to explore the scheduling of assisted and unassisted practice over the course of a training regimen. In future work, we plan to test subjects in higher-dimensional tasks and make comparisons to other assist-as-needed controllers, such as path controllers, active constraints, and other impedance-based approaches. In addition, we are exploring ways to define more complex tasks where it may be difficult to define a desired trajectory or goal state.

\section{Conclusion}

Numerous devices and control strategies have been developed to facilitate forceful interaction between humans and robots for the purposes of training specific skills or tasks. However, it is difficult to show the efficacy of these robots in promoting skill learning. Some types of robot-mediated training may be detrimental to learning, and others might be no more effective than an equivalent amount of unassisted practice. Interfaces for pHRI that have been shown to successfully enhance training have several features explicitly included in their design to enhance motor learning. Specifically, the automation must be able to assist users in completing the task and adapt the assistance to the needs of the individual user in terms of both initial skill and current performance in order to promote user engagement.

In this work, we investigate the use of a hybrid shared control method for assistance and training. The interface allowed subjects to make errors and even fail at the task. While the application of the filter improved subject success rates, it did not make subjects successful all of the time. It also avoided enforcing a specific trajectory by evaluating the effect of user inputs on a continuous basis. Results from two user studies with different task-based acceptance criteria demonstrate the method's effectiveness in both assistance and training. Analysis of the correlations between the level of controller engagement and the initial skill of the users showed that the filter is sensitive to users' skill level. While the filter inherently adapts with every measurement of the user inputs, the strong correlation between performance measures and the level of controller intervention shows that this instantaneous adaptation results in a controller that also assists as needed according to the performance of the user in an individual trial. 

\begin{funding}
This work was supported by the National Science Foundation under grant 1637764 and by the Department of Defense (DoD) through the National Defense Science \& Engineering Graduate Fellowship (NDSEG) Program. Any opinions, findings, and conclusions or recommendations expressed in this material are those of the authors and
do not necessarily reflect the views of the National Science
Foundation or of the NDSEG program.
\end{funding}

\begin{acks}
The authors would like to thank Sabeen Admani for her unwavering support in debugging the robot and keeping our experiments on schedule.
\end{acks}

\theendnotes

\section*{Supplementary Materials}
The experimental data presented in this article can be found online by following the hyperlinks from www.ijrr.org.

\begin{table}[h!]
\begin{tabular}{l p{6.2cm}}
\multicolumn{2}{l}{\textbf{\sffamily Table of Experimental Data}}\\
\hline
\textbf{Data File} & Description\\ \hline
\href{https://murpheylab.github.io/data/2019IJRRFiKaDeMu-S1.csv}{\textcolor{blue}{S1}} & Performance metrics calculated for each trial and participant session in the OCIP study.\\
\href{https://murpheylab.github.io/data/2019IJRRFiKaDeMu-S2.csv}{\textcolor{blue}{S2}} &  An example of trajectories collected from a single participant in their first session without assistance in the OCIP study. \\
\href{https://murpheylab.github.io/data/2019IJRRFiKaDeMu-S3.csv}{\textcolor{blue}{S3}} & An example of trajectories collected from a single participant in their second session with assistance in the OCIP study.\\
\href{https://murpheylab.github.io/data/2019IJRRFiKaDeMu-S4.csv}{\textcolor{blue}{S4}} & Performance metrics calculated for each trial and participant set in the MIG study.\\
\href{https://murpheylab.github.io/data/2019IJRRFiKaDeMu-S5.csv}{\textcolor{blue}{S5}} &  An example of trajectories collected from a single participant in their first set without assistance in the MIG study. \\
\href{https://murpheylab.github.io/data/2019IJRRFiKaDeMu-S6.csv}{\textcolor{blue}{S6}} &  An example of trajectories collected from a single participant in their second set with assistance in the MIG study. \\
\href{https://murpheylab.github.io/data/2019IJRRFiKaDeMu-S7.csv}{\textcolor{blue}{S7}} &  An example of trajectories collected from a single participant in their third set without assistance in the MIG study. \\
 \hline
\end{tabular}
\end{table}
\end{document}